\pdfoutput=1
\documentclass[conference]{IEEEtran}
\IEEEoverridecommandlockouts
\usepackage{amsmath,amssymb,amsfonts}
\usepackage{algorithmic}
\usepackage{graphicx}
\usepackage{textcomp}
\usepackage{xcolor}
\usepackage{multicol}
\usepackage{comment}
\usepackage{booktabs, multirow} 
\usepackage{enumitem}
\definecolor{codegreen}{rgb}{0,0.6,0}
\definecolor{codegray}{rgb}{0.5,0.5,0.5}
\definecolor{codepurple}{rgb}{0.58,0,0.82}
\definecolor{backcolour}{rgb}{0.95,0.95,0.92}
\usepackage[justification=centering]{caption}
\usepackage{listings}
\usepackage{makecell}
\usepackage{subcaption}
\usepackage{threeparttable}
\usepackage{biblatex}
\lstdefinestyle{mystyle}{
    backgroundcolor=\color{backcolour},   
    commentstyle=\color{codegreen},
    keywordstyle=\color{magenta},
    numberstyle=\tiny\color{codegray},
    stringstyle=\color{codepurple},
    basicstyle=\ttfamily\scriptsize,
    breakatwhitespace=false,         
    breaklines=true,                 
    captionpos=t,                    
    keepspaces=true,                 
    numbersep=5pt,                  
    showspaces=false,                
    showstringspaces=false,
    showtabs=false,                  
    tabsize=2,
}

\def\BibTeX{{\rm B\kern-.05em{\sc i\kern-.025em b}\kern-.08em
    T\kern-.1667em\lower.7ex\hbox{E}\kern-.125emX}}

\begin{document}

\title{Making Machine Learning Datasets and Models FAIR for HPC: A Methodology and Case Study
\thanks{This work is supported by the U.S. Department of Energy, Office of Science, Advanced Scientific Computing Program under contract number DE-AC02-06CH11357 and Award Number DE-SC0021293. Prepared by LLNL under Contract DE-AC52-07NA27344 (LLNL-CONF-832022).}
}

\author{\IEEEauthorblockN{ 
  Pei-Hung Lin\IEEEauthorrefmark{1},
  Chunhua Liao\IEEEauthorrefmark{1},
  Winson Chen\IEEEauthorrefmark{1}\IEEEauthorrefmark{4}, 
  Tristan Vanderbruggen\IEEEauthorrefmark{1},  \\ 
  Murali Emani\IEEEauthorrefmark{3},
  Hailu Xu\IEEEauthorrefmark{2}
  }
\IEEEauthorrefmark{1}Lawrence Livermore National Laboratory, Livermore, CA 94550, USA \\
\IEEEauthorrefmark{2}California State University, Long Beach , CA 90840, USA\\
\IEEEauthorrefmark{3}Argonne National Laboratory, Lemont, IL 60439, USA \\
\IEEEauthorrefmark{4}University of California Santa Cruz, Santa Cruz, CA 95064, USA\\
}

\maketitle

\begin{abstract}
The FAIR Guiding Principles aim to improve the findability, accessibility, interoperability, and reusability of digital content by making them both human and machine actionable.
However, these principles have not yet been broadly adopted in the domain of machine learning-based program analyses and optimizations for High-Performance Computing (HPC).
In this paper, we design a methodology to make HPC datasets and machine learning models FAIR after investigating existing FAIRness assessment and improvement techniques. Our methodology includes a comprehensive, quantitative assessment for elected data, followed by concrete, actionable suggestions to improve FAIRness with respect to common issues related to persistent identifiers, rich metadata descriptions, license and provenance information.
Moreover, we select a representative training dataset to evaluate our methodology. 
The experiment shows the methodology can effectively improve the dataset and model's FAIRness from an initial score of 19.1\% to the final score of 83.0\%.
\end{abstract}


\section{Introduction} \label{introduction}
Research activities in high-performance computing (HPC) community have applied machine learning (ML) for various research needs such as performance modeling and prediction~\cite{lopez2014survey}, 
memory optimization~\cite{xu2019machine, xunified}, and so on. %
A typical ML-enabled HPC study generates a large amount of valuable datasets from the HPC experiment outputs. The datasets serve as training inputs for the ML models applied in research. 
There is an increasing awareness of the need for data reuse in HPC community.
However, the community still lacks guidelines and experiences to effectively access and share the data that was collected in various research experiments.

The FAIR Guiding Principles~\cite{wilkinson2016fair} were published to advocate the reuse of data and other digital contents, including algorithms, software tools, source codes, and workflows that led to the generation of data.
These principles serve as a guideline, but not specification, to make digital data and contents findable, accessible, interoperable, and reusable by both humans and machines.
The FAIR principles have been broadly adopted in certain research communities, such as biomedical and health science, with a range of studies and implementations to address the challenges in achieving data FAIRness.
Many recommendations and best practices are documented to prepare FAIR data in specific research domains.
There are also various initiatives and active community activities, focusing the generalist, started to promote the importance of data FAIRness and assist to FAIRify the existing datasets. 
However, these principles and practices haven’t yet been tailored to benefit the HPC community. 

In this paper, we present a methodology and a case study to make HPC machine learning training datasets FAIR, in the domain of machine learning-based program analyses and optimizations for HPC.
We investigate the needs and challenges to FAIRify a HPC dataset following the established data FAIRification workflows used in other research communities.
Seeing limitations of existing FAIRness evaluation approaches, we propose a hybrid assessment step with both manual and automated assessments. Concrete actions are then suggested to address missing features in the chosen dataset. The methodology is evaluated using a representative dataset, demonstrating significant FAIRness improvement from 19.1\% to 83.0\%. 

This paper has the following contributions:
\begin{itemize}
    \item We survey the best practices and approaches to make datasets FAIR, including those for FAIRness assessment and improvements (or FAIRification). 
    \item A comprehensive, quantitative methodology is proposed to evaluate and improve FAIRness of HPC datasets. 
    \item The methodology leverages a hybrid assessment to generate a single FAIRness score. An ontology is developed to help address common issues for FAIRness improvement. 
    \item Using a concrete dataset, we evaluate the proposed methodology for its effectiveness.
\end{itemize}

The remainder of this paper is organized as follows: Section \ref{FAIR} gives background information about the FAIR principles. 
Section \ref{Fairify} discusses the methodology we design to evaluate and improve FAIRness of HPC datasets and ML models.
Section \ref{dataset} presents an example dataset and an example ML model generated from a prior project named XPlacer \cite{xu2019machine}. Experiment results are shown in Section~\ref{results}.  Section \ref{relatedwork} summarizes the related work and section \ref{conclusion} concludes this paper.

\section{Background of the FAIR Guiding Principles}\label{FAIR}
\textbf{FAIR Guiding Principles: }
With a massive amount of data generated and collected in research activities, the scientific communities are pursuing good data management and data stewardship to have high quality digital publications that can simplify data discovery, evaluation, and reuse in downstream studies.  
The FAIR Guiding Principles, standing for findable, accessible, interoperable and reusable,  were jointly prepared by representatives and interested stakeholder groups
as guidelines for data management and stewardship~\cite{wilkinson2016fair}. 

\noindent\textbf{FAIRness Evaluation:}
FAIRness evaluation plays an important role by providing assessments to assist data generators and stewards to improve the data FAIRness. 
The FAIR data maturity model released by the Research Data Alliance (RDA)\cite{rda2020fair} defines three elements in an evaluation framework: 1) FAIRness indicators derived from the FAIR principles to formulate measurable aspects of each principle; 2) priorities reflecting the relative importance of the indicators; and 3) the evaluation method defining a quantitative approach to report the evaluation results.
Currently, three major types of FAIRness evaluation are available to various research communities: 
\begin{enumerate}[leftmargin=*]

\item{\textit{Discrete-answer questionnaire-based evaluation}}: This approach provides a checklist of single-selection questions 
to reflect the FAIR principles and related concepts. 
    It is usually accompanied by a scoring system to evaluate the FAIRness.  It requires little knowledge about the FAIR principles and is relatively straightforward to exploit the evaluation with discrete-answer questionnaire.  
    An example question reflecting the 'F1' FAIR principle would ask if the data has persistent identification (PID), metadata and documentation describing the data.
    A zero point score can be given if no PID, metadata and/or documentation is available.  Whereas a non-zero score can be assigned for other possible answers. 
    An example of the discrete-answer evaluation is the FAIRness evaluation tool provided from RDA FAIR data maturity model. 
    
    \item{\textit{Open-answer questionnaire-based evaluation}:} 
    This approach also exploits a list of metrics reflecting the FAIR principles. 
    Different from the discrete-answer approach, the open-answer approach requires concrete answers and statements to the metric as evidences to the implementation of FAIRness.  
    An example metric to the 'F1' FAIR principle is: Whether the Globally Unique Identifier (GUID) matches a GUID scheme recognized as being globally unique in a global public registry.  The evaluation expects a valid GUID submitted as an answer. 
    The open-answer approach allows the scientific communities to extend the evaluation by creating additional metrics. However, it shares the  disadvantages as the discrete-answer approach that the manual process to input answers is required and the possible biased results. 
    
    
    \item{\textit{Automated evaluation}}: This approach automatically retrieves and evaluates a given digital resource, simplifies the evaluation process and eliminates the human bias. 
    Automated evaluation service could also provide evaluation feedback and recommendation to improvement. 
    However, the evaluation capability of this approach is limited by the metrics chosen, the software support, target granularity of data objects, and the resource availability of the metadata providers.  For example,
    two evaluation tools,
    F-UJI\footnote{https://www.f-uji.net/index.php} 
    and FAIR-checker\footnote{https://fair-
checker.france-bioinformatique.fr/base metrics}, use different number of metrics to evaluation the license information for 'R' FAIR principle.
     Existing automated FAIR assessment services only inspect the metadata information at a coarse granularity. They check the descriptive metadata for the whole dataset and inspect the format of the dataset but do not inspect the FAIRness of data elements, the contents of the dataset.  

\end{enumerate}

\noindent\textbf{FAIRification Processes:}
As the FAIR Guiding Principles get higher attention in various research communities, many initiatives started to advocate the application of FAIR data principles.  GO FAIR\cite{GOFAIR}, a global, stakeholder-driven and self-governed initiative, has been working towards implementations of the FAIR Guiding Principles and proposes FAIRification framework with three focuses: 
First, the formation of machine-actionable metadata; Second, the collection of machine-actionable FAIR implementation profiles (FIPs); And last, a global FAIR data service.
This framework  is designed to serve as the “how to” guidance to stakeholders seeking to go FAIR. 
It exploits a bottom-up approach to start with metadata preparations for datasets from various communities.  The collection of FIPs enlarges the metadata coverage to be shared among communities. 
Ultimately, a global FAIR data service stores and hosts information about datasets to provide data Accessibility, Findability and Reusability.

\begin{figure*}[!htbp]
 \centering
 \includegraphics[width =1.0\textwidth]{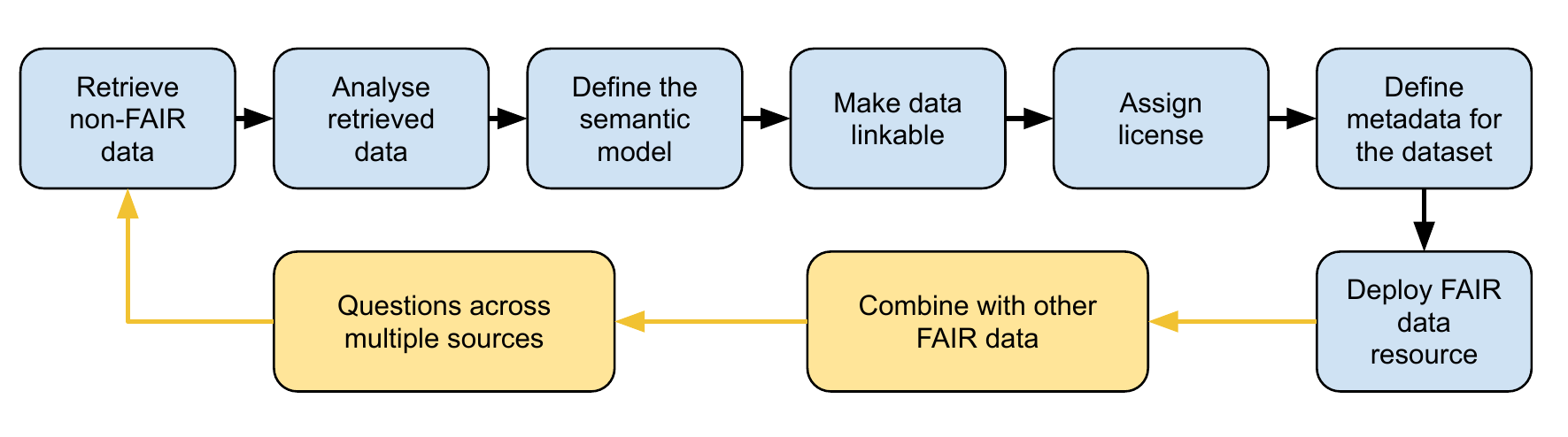}
 \caption{\textmd{FAIRification Process proposed by GO FAIR initiative.}}
 \label{fig:FAIRification}
\end{figure*}

Together with the FAIRification framework, the GO FAIR initiative also provides a FAIRification process aiming at addressing the conversion of raw datasets into FAIR datasets. 
These 7 steps and the the workflow are presented in Figure~\ref{fig:FAIRification}. 
This FAIRification process emphasizes on FAIRification for both data and metadata, including seven steps: (1) retrieve non-FAIR data, (2) analyze the retrieved data, (3) define the semantic model, (4) make data linkable, (5) assign license, (6) define metadata for the dataset, and (7) deploy/publish FAIR data resource. 
The process is proposed for general purpose without considering specific requirements and needs from individual community.  
Challenges may arise when applying the process for data coming from a specific domain.
For instance, without standardized data format in HPC community, the analyses used for the retrieved data vary based on the selected custom data formats.
The development of a HPC semantic model needs iterative reviews and revisions to ensure consistency in providing accurate and unambiguous meaning of entities and relations
in the HPC domain.


\section{A Methodology to Make Datasets FAIR}
\label{Fairify}

We propose a new methodology to improve the FAIRness of HPC datasets. The method has three steps: 1) initial FAIRness assessment, 2) improving FAIRness based on assessment results, and 3) final assessment. We elaborate the first two steps since the last step is a repetition of the first one.  

\subsection{Initial FAIRness Assessment}
After surveying the existing FAIRness evaluation methods, we have found that while
the questionnaire-based manual evaluations have advantages to cover more details, the manual input process and the tendency to cause biased result are not ideal (shown in Table~\ref{tab:manualVsAuto}).
On the other hand, automated evaluation avoids manual intervention and biased result.  But they are less flexible and limited by the implementation for supported data granularity and metadata. 
As a result, we propose a hybrid approach involving both manual and automated assessments to have a more productive, flexible and informative evaluation.

We choose the FAIRness assessment service by F-UJI \cite{DEVARAJU2021100370} as the automated evaluation service due to many desirable features it has.
The F-UJI service designs 17 FAIRness metrics, derived from the 41 RDA FAIRness maturity indicators, that can be evaluated with automatic testings. 
It uses a scoring system to provide quantitative evaluation result, in percentage,  reflecting the ratio of the achieved score number to the full score number. 
Moreover,  F-UJI provides detailed output to each metric evaluation as a guidance to assist users to improve the FAIRness maturity. Finally, F-UJI is an open-source project that can be expanded to include new features and support to address specific needs for HPC community.
However,  F-UJI has insufficient support for many FAIR Guiding Principals related to metadata. 

\begin{table*}[!htbp]
	\centering
\begin{tabular}{|l|l|l|l|} \hline
\textbf{Criteria} & \textbf{Description} &\textbf{Manual}  & \textbf{Automated} \\ \hline
Metadata  & Metadata information that can be evaluated & Flexible  & Fixed \\  \hline
Productivity & Human effort required for evaluation & Low  &  High \\ \hline
Completeness & FAIR Guiding Principles coverage & High & Low 	\\ \hline
Granularity & Data granularity that can be evaluated & Yes &  No \\ \hline
\end{tabular}
\caption{Comparison of manual and automated assessment.
}
\label{tab:manualVsAuto}
\end{table*}

The self-assessment tool by RDA FAIR data maturity model is chosen as the manual evaluation approach. It uses the same 41 RDA FAIRness maturity indicators referenced by F-UJI.
This manual assessment is a discrete-answer questionnaire-based evaluation. Users provide an answer, with an associated level number, to each indicator from one of the following options to determine the maturity: (0) not applicable; (1) not been considered; (2) under consideration or in planning; (3) in implementation; and (4) fully implemented.  Visualized report is available from the tool to present the maturity of the FAIRness. The manual evaluation provides good coverage for the details in the FAIR principles and can present the full details of the FAIRness readiness. 
It also provides flexibility to evaluate FAIRness of data of different granularity, ranging from metadata, datasets, to data elements. However, manually checking each indicator to evaluate given data is tedious and error-prone. 

The RDA indicators and F-UJI metrics share the same notation with three components to name their IDs:
(1) the project abbreviation, RDA or FsF (FAIRsFAIR: the parent project of F-UJI); (2) the related FAIR Guiding Principle identifier (e.\,g. F1); and (3) a local identifier number with a suffix encoding to identify resource that will be evaluated: metadata (M), data (D), or both (MD).  

In the proposed hybrid assessment, we consolidate 41 RDA indicators and 17 F-UJI metrics to form a new list of 47 indicators as shown in Table~\ref{tab:fairness}.
There are 11 indicators
, marked $\ast$ in Table~\ref{tab:fairness}, 
originated from both RDA indicators and F-UJI metrics, e.\,g. RDA-F1-01D and FsF-F1-02D. 
The hybrid evaluation for these 11 indicators will take the result from the F-UJI metric because the automated test has the advantage to avoid biased result. 

To generate quantitative assessment results, we design a scoring system as follows: each indicator is given one point if its result is determined by the RDA maturity indicator as fully implemented, or by the F-UJI metric as a fully passed test. A total of 47 points for the whole hybrid evaluation consist 8 points for 'F', 13 points for 'A', 14 points for 'I' and 12 points for 'R', where the points correspond to the number of FAIRness indicators for each FAIR principle.
The final score is represented as the percentage of the earned points divided by the total point count. 




\begin{table*}[!htp]\centering
\caption{FAIR Guiding Principles: maturity indicators and assessment metrics}
\label{tab:fairness}
\scriptsize
\begin{threeparttable}[b]
\begin{tabular}{|l|l|p{10cm}|c|} \hline
\textbf{P\tnote{1}}& \textbf{Indicator} &\textbf{Description} & \textbf{S\tnote{2}} \\ \hline

\multirow{6}{*}{\hfil F1} & RDA-F1-01M & Metadata is identified by a persistent identifier & \multirow{6}{*}{\hfil 4}\\
\cline{2-3}

& RDA-F1-01D \tnote{*} & \multirow{2}{*}{Data is identified by a persistent identifier} & \\

& FsF-F1-02D & & \\
\cline{2-3}

& RDA-F1-02M & Metadata is identified by a globally unique identifier & \\
\cline{2-3}

& RDA-F1-02D \tnote{*}& \multirow{2}{*} {Data is identified by a globally unique identifier} & \\

& FsF-F1-01D & &\\
\hline
\multirow{2}{*}{\hfil F2} & RDA-F2-01M &  Rich metadata is provided to allow discovery & \multirow{2}{*}{\hfil 2} \\
\cline{2-3}

& FsF-F2-01M & Metadata includes descriptive core elements to support data findability &\\
\hline

\multirow{2}{*}{\hfil F3} & RDA-F3-01M \tnote{*}&  \multirow{2}{*}{Metadata includes the identifier for the data} & \multirow{2}{*}{\hfil 1} \\
& FsF-F3-01M & & \\
\hline

\multirow{2}{*}{\hfil F4}  & RDA-F4-01M \tnote{*}&  \multirow{2}{*}{Metadata is offered in such a way that it can be harvested and indexed} & \multirow{2}{*}{\hfil 1} \\

&FsF-F4-01M & &\\
\hline
\hline

\multirow{11}{*}{\hfil A1} & RDA-A1-01M &  Metadata contains information to enable the user to get access to the data & \multirow{11}{*}{\hfil 9}\\

\cline{2-3}

 & RDA-A1-02M &  Metadata can be accessed manually &\\
 
\cline{2-3}

& RDA-A1-02D &  Data can be accessed manually & \\
 
\cline{2-3}

 & RDA-A1-03M &  Metadata identifier resolves to a metadata record or digital object & \\

\cline{2-3}

& RDA-A1-03D &  Data identifier resolves to a metadata record or digital object  &\\

\cline{2-3}
 & RDA-A1-04M \tnote{*}& \multirow{2}{*}{Metadata is accessed through standardised protocol}& \\
&FsF-A1-02M && \\

\cline{2-3}
 & RDA-A1-04D \tnote{*}&  \multirow{2}{*}{Data is accessed through standardized protocol} &\\
& FsF-A1-03D &  &\\
\cline{2-3}

& RDA-A1-05D &  Data can be accessed automatically &\\

\cline{2-3}
& FsF-A1-01M & Metadata contains access level and access conditions of the data &\\

\hline
\multirow{2}{*}{\hfil A1.1} & RDA-A1.1-01M &  Metadata is accessible through a free access protocol &\multirow{2}{*}{\hfil 2} \\

\cline{2-3}

 & RDA-A1.1-01D &  Data is accessible through a free access protocol & \\

\hline

A1.2 & RDA-A1.2-01D &  Data is accessible through an access protocol that supports authentication and authorization & \multirow{2}{*}{\hfil 1} \\
\hline

\multirow{2}{*}{\hfil A2}  & RDA-A2-01M \tnote{*} &  \multirow{2}{*}{Metadata is guaranteed to remain available after data is no longer available} & \multirow{2}{*}{\hfil 1}\\

&FsF-A2-01M& &\\
\hline
\hline

\multirow{6}{*}{\hfil I1} & RDA-I1-01M &  Metadata uses knowledge representation expressed in standardized format & \multirow{6}{*}{\hfil 6}\\

\cline{2-3}

& RDA-I1-01D &  (Data uses knowledge representation expressed in standardized format &\\

\cline{2-3}
 & RDA-I1-02M &  Metadata uses machine-understandable knowledge representation  &\\
 
\cline{2-3}
 & RDA-I1-02D &  Data uses machine-understandable knowledge representation  &\\
 
 \cline{2-3}
 &FsF-I1-01M & Metadata is represented using a formal knowledge representation language &\\
 \cline{2-3}
 &FsF-I1-02M & Metadata uses semantic resources &\\
 \hline
 
 \multirow{2}{*}{\hfil I2}  & RDA-I2-01M &  Metadata uses FAIR-compliant vocabularies   & \multirow{2}{*}{\hfil 2}\\

 \cline{2-3}

 & RDA-I2-01D &  Data uses FAIR-compliant vocabularies & \\
 \hline

\multirow{7}{*}{\hfil I3} & RDA-I3-01M \tnote{*}&  \multirow{2}{*}{Metadata includes references to other (meta)data} & \multirow{7}{*}{\hfil 6}\\

&FsF-I3-01M& & \\
\cline{2-3}

& RDA-I3-01D &  Data includes references to other (meta)data & \\
\cline{2-3}
& RDA-I3-02M &  Metadata includes references to other data & \\

\cline{2-3}
& RDA-I3-02D &  Data includes references to other data &  \\

\cline{2-3}

& RDA-I3-03M &  Metadata includes qualified references to other metadata  & \\

\cline{2-3}

& RDA-I3-04M &  Metadata include qualified references to other data & \\

\hline
\hline

\multirow{3}{*}{\hfil R1} & RDA-R1-01M &  Plurality of accurate and relevant attributes are provided to allow reuse & \multirow{2}{*}{\hfil 2} \\
\cline{2-3}

&FsF-R1-01MD & Metadata specifies the content of the data & \\
\hline

\multirow{4}{*}{\hfil R1.1} & RDA-R1.1-01M & Metadata includes information about the license under which the data can be reused & \multirow{4}{*}{\hfil 3}\\

\cline{2-3}
& RDA-R1.1-02M &  Metadata refers to a standard reuse license & \\
\cline{2-3}
 & RDA-R1.1-03M \tnote{*}&  \multirow{2}{*}{Metadata refers to a machine-understandable reuse license} & \\
& FsF-R1.1-01M & & \\
\hline
\multirow{3}{*}{\hfil R1.2} & RDA-R1.2-01M &  Metadata includes provenance information according to community-specific standards & \multirow{3}{*}{\hfil 3} \\

\cline{2-3}

& RDA-R1.2-02M &  Metadata includes provenance information according to a cross-community language & \\

\cline{2-3}

&FsF-R1.2-01M & Metadata includes provenance information about data creation or generation & \\
\hline

\multirow{6}{*}{\hfil R1.3} & RDA-R1.3-01M \tnote{*}&  \multirow{2}{*}{Metadata complies with a community standard } & \multirow{6}{*}{\hfil 4}\\
&FsF-R1.3-01M & &\\
\cline{2-3}
& RDA-R1.3-01D &  Data complies with a community standard & \\

\cline{2-3}

& RDA-R1.3-02M &  Metadata is expressed in compliance with a machine-understandable community standard & \\
\cline{2-3}

& RDA-R1.3-02D \tnote{*}&  \multirow{2}{*}{Data is in compliance with a machine-understandable community standard} &\\
&FsF-R1.3-02D & &\\

\hline

\end{tabular}
\begin{tablenotes}
    \item[1] FAIR Guiding Principle ID
    \item[2] Max score allocated to each sub-principle
    \item[*] Both RDA FAIRness indicator and F-UJI metric represent the same evaluation
  \end{tablenotes}
 \end{threeparttable}

\end{table*}

\subsection{Improving FAIRness}
\label{sec:improvingMethods}
Improving the FAIRness can be achieved by systematically addressing issues reported by the manual and automated assessments.
Users can iterate through all the metrics that are not marked as fully implemented or passed. The evaluation reports often give reasons for the failed tests associated with the metrics. 
Due to page limit, we present example actions users can take to address some commonly seen inadequacies in FAIRness revealed by our hybrid assessment process. 
We also label these actions by identifiers with a prefix MT (abbreviation for method). 

\begin{itemize}[leftmargin=*]
\item \textit{Getting persistent identifier} (MT-PID):
It is straightforward for a data collector to use the web address (URL) as the identifier for a dataset.  However, URLs tend to change over time which leads to broken links to the data. 
A persistent and unique identifier, such as digital object identifier (DOI), is the preferred identifier for a FAIR dataset.
It is recommended to register the dataset at general or domain-specific registry systems to make the data more discoverable.
We select Zenodo\footnote{https://zenodo.org/}, a general-purpose open access repository, as our default platform to generate DOIs. 

\item \textit{Providing coarse-grain metadata information} (MT-coarse-metadata):
Inadequate metadata is a common issue for datasets to fulfill the FAIR principles.
Again, we use public data hosting services such as \texttt{zenodo.org} to leverage their  built-in metadata.  The metadata information are provided by filling in the required information during the data registering/uploading process.
This type of metadata tends to describe general information of the associated dataset.
Data collectors can also prepare metadata by following guidance for general data, e.\,g. Dublin Core metadata initiative, or domain-specific data, e.\,g. Data Documentation Initiative (DDI) for social, behavioral, economic and health sciences.
With manual FAIRness evaluation, users have greater flexibility to prepare the metadata.  Specific metadata formats and requirements should be fulfilled if the dataset is assessed by an automated evaluation service.
The F-UJI automated evaluation supports several multidisciplinary meatadata standards, e.\,g. Dublin Core and DataCite Metadata Schema, and metadata standards from several scientific domains, e.\,g. biology, botany, and paleontology.  General metadata information supported by \texttt{zenodo.org} are recognized by the F-UJI evaluation. 

\item \textit{Generating rich attributes for different granularity of data} (MT-fine-metadata). It is relatively easy to make a whole dataset FAIR. However, significant work is needed to make fine-grain data elements inside a dataset FAIR. There is a lack of community standards to provide rich attributes for the data elements of various science domains. In HPC, datasets can be generated from performance profiling tools, compilers, runtime systems with very different attributes. 
To address this problem, we are developing the HPC Ontology~\cite{liao2021hpc} to provide standard attributes which can be used to annotate fine-grain data elements in different subdomains of HPC, including GPU profiling results, program analysis (e.\,g. call graph analysis), and machine learning models (e.\,g. decision trees). The design of HPC Ontology is modular so it can be extended to include more subdomains in the future. 

\item \textit{Automatic annotating data elements} (MT-auto-annotating). Even with existing standard attributes, it is impractical to manually annotate data elements one by one. 
Since most datasets are published in CSV files, we leverage Tarql\footnote{https://tarql.github.io/}, a command-line tool for converting data stored in CSV files into the Resource Description Framework (RDF) format using standard metadata and attributes provided by the HPC Ontology. 

\item \textit{Provenance information} (MT-provenance):
Provenance information is required to fulfill `R' FAIR principle.
Similar to the metadata, basic provenance information such as publication date and publisher, can be provided through the data hosting service such as zenodo.org.  
Formal provenance metadata, such as PROV, is recommended for advanced provenance information support.

\item \textit{License information} (MT-license):
  Data collectors should choose and apply license information to the collected dataset to fulfill `R1.1' FAIR principle. A recommended choice for data is one of the Creative Commons license. 

\end{itemize}

\begin{table*}[!htp]\centering
\caption{Examples of HPC Ontology properties to annotate fine-grain GPU profiling data elements}
\label{tab:xplacer-properties}
\scriptsize
\begin{tabular}{|l|l|l|}\hline
\textbf{Property} &\textbf{Data type} &\textbf{Description} \\\hline
hpc:benchmark & xsd:anyURI & Link to the associated benchmark software ID in ontology\\
hpc:kenelName & xsd:string & Kernel name\\
hpc:gpuThreadBlockSize & xsd:integer & Launch block size \\
hpc:registersPerThread & xsd:integer & Register usage per thread \\
hpc:gpuThreadCount & xsd:integer & Threads count in launched kernel \\
hpc:gpuWavesPerSM & xsd:integer & Wave count in SM \\
hpc:maxGPUThreadBlockSizeLimitedByRegister & xsd:integer & Max block limited by registers \\
hpc:maxGPUThreadBlockSizeLimitedByWarps & xsd:integer & Max block limited by warps \\
hpc:cpuPageFault & xsd:integer &  CPU page fault count\\
hpc:gpuPageFault & xsd:integer &  GPU page fault count\\
hpc:hostToDeviceTransferSize & qudt:QuantityValue & Host to Device data transfer size \\
hpc:deviceToHostTransferSize & qudt:QuantityValue & Device to Host data transfer size\\

\hline
\end{tabular}
\end{table*}

\todo{We should include some details about how we develop metadata (HPC Ontology), how to annotate existing datasets with metadata, etc.  Again , a table listing principles/metrics and how to improve the relevant metrics is best with concrete steps/URLs readers can follow.}

\section{HPC Ontology}
\label{HPCOntology}
     An  ontology is a formal specification for explicitly representing knowledge about types,  properties, and interrelationships of the  entities in a domain. 
     An ontology can provide the much needed controlled vocabularies, knowledge representation, and standards to fulfill Interoperability ("I") and  Reusability ("R")  of  the  FAIR data principles.   
     
     We have developed the HPC ontology to cover HPC specific properties including software, hardware, and their interactions.  
     The coverage of HPC ontology is expected to expand gradually as more data and information are explored.
     HPC ontology is represented with JSON-LD, a method of encoding linked data using JSON format,  for expressing data in the Resource Description Framework (RDF) data model.
     HPC ontology is described in Terse RDF Triple Language, or Turtle, for expressing data in the Resource Description Framework (RDF) data model.
     A two-level design is adopted for HPC ontology to effectively cover both high-level concepts needed in the HPC domain and more specific low-level details in subdomains. 
     The high-level components  of the HPC ontology aim to  provide descriptive,  administrative  and  statistical metadata  for  users  to annotate  datasets. 
     The top level concept, \texttt{hpc:Thing}, in the HPC ontology describes a set of fundamental properties including its unified resource identifier, the type of the ID , name, url, etc. 
     All other high-level concepts are directly or indirectly subclasses of \texttt{hpc:Thing}. Each concept inherits the fundamental properties from its superclass and also includes additional properties as needed.
     
     On the other hand, the set of  the low-level components provides concepts and relationships details for smaller domains and aims to be extensible for support rapid changes in HPC software and hardware.
     They are essential in HPC ontology by providing rich attributes to describe data elements to achieve maximal FAIRness.
     The low-level components in HPC ontology will relevant scopes in HPC including hardware, software, performance data, and AI models.
     Table~\ref{tab:xplacer-properties} list examples of properties defined in HPC ontology and needed by the Xplacer dataset.

\begin{table*}[!htp]\centering
\caption{Examples of HPC Ontology properties to annotate Xplacer dataset}
\label{tab:xplacer-properties}
\scriptsize
\begin{tabular}{|l|l|l|}\hline
\textbf{Property} &\textbf{Data type} &\textbf{Description} \\\hline
hpc:benchmark & xsd:anyURI & Link to the associated benchmark software ID in ontology\\
hpc:kenelName & xsd:string & Kernel name\\
hpc:gpuThreadBlockSize & xsd:integer & Launch block size \\
hpc:registersPerThread & xsd:integer & Register usage per thread \\
hpc:gpuThreadCount & xsd:integer & Threads count in launched kernel \\
hpc:gpuWavesPerSM & xsd:integer & Wave count in SM \\
hpc:maxGPUThreadBlockSizeLimitedByRegister & xsd:integer & Max block limited by registers \\
hpc:maxGPUThreadBlockSizeLimitedByWarps & xsd:integer & Max block limited by warps \\
hpc:cpuPageFault & xsd:integer &  CPU page fault count\\
hpc:gpuPageFault & xsd:integer &  GPU page fault count\\
hpc:hostToDeviceTransferSize & qudt:QuantityValue & Host to Device data transfer size \\
hpc:deviceToHostTransferSize & qudt:QuantityValue & Device to Host data transfer size\\

\hline
\end{tabular}
\end{table*}

\section{XPlacer Datasets}
\label{dataset}
In this paper, we pick a dataset and ML model generated by the XPlacer~\cite{xu2019machine} to study how to FAIRify HPC datasets and ML models. 
The reason to choose XPlacer is that the authors released raw data with detailed documents explaining how data was generated and processed. 
They also describe the meanings of each row and column for the corresponding CSV files.
The rich human-readable documentation provides a good foundation for FAIRness evaluation and improvement. 

XPlacer is a memory optimization tool developed to use machine learning to guide the optimal use of memory APIs available on Nvidia GPUs. 
Programmers for GPUs traditionally use a set of APIs to explicitly move data between CPU and GPU memory. 
Unified memory (UM), introduced in modern Nvidia GPU systems, greatly reduces GPU programming burdens by providing hardware-managed data migration.
A set of memory advisory APIs is also provided by Nvidia to advise the UM driver to improve performance. 
The motivation is that Nvidia provides a rich set of memory APIs and corresponding parameters. It is challenging for programmers to manually decide which API to use and with what parameter values. XPlacer aims to build machine learning models to automatically make such decisions for a program running on heterogeneous CPU/GPU machines.
It has both offline training and online adaptation steps, including collecting training datasets, generating machine learning models, and applying the generated model to predict the best memory usage advise.


The offline data collection uses seven benchmarks from the Rodinia benchmark~\cite{che2009rodinia} with seven available memory advises applied  to different arrays in the benchmarks.
In each experiment with a selected benchmark, profiling tools were used to collect kernel level and data object level metrics.
A feature correlation algorithm is applied to down select the 10 most relevant features from the original 59 metrics gathered in the metric collection.
After data normalization and feature dimensionality reduction, the collected XPlacer dataset has a total of 2688 samples prepared for the machine learning models.

\begin{figure}[!ht]
 \centering
 \includegraphics[scale=0.45]{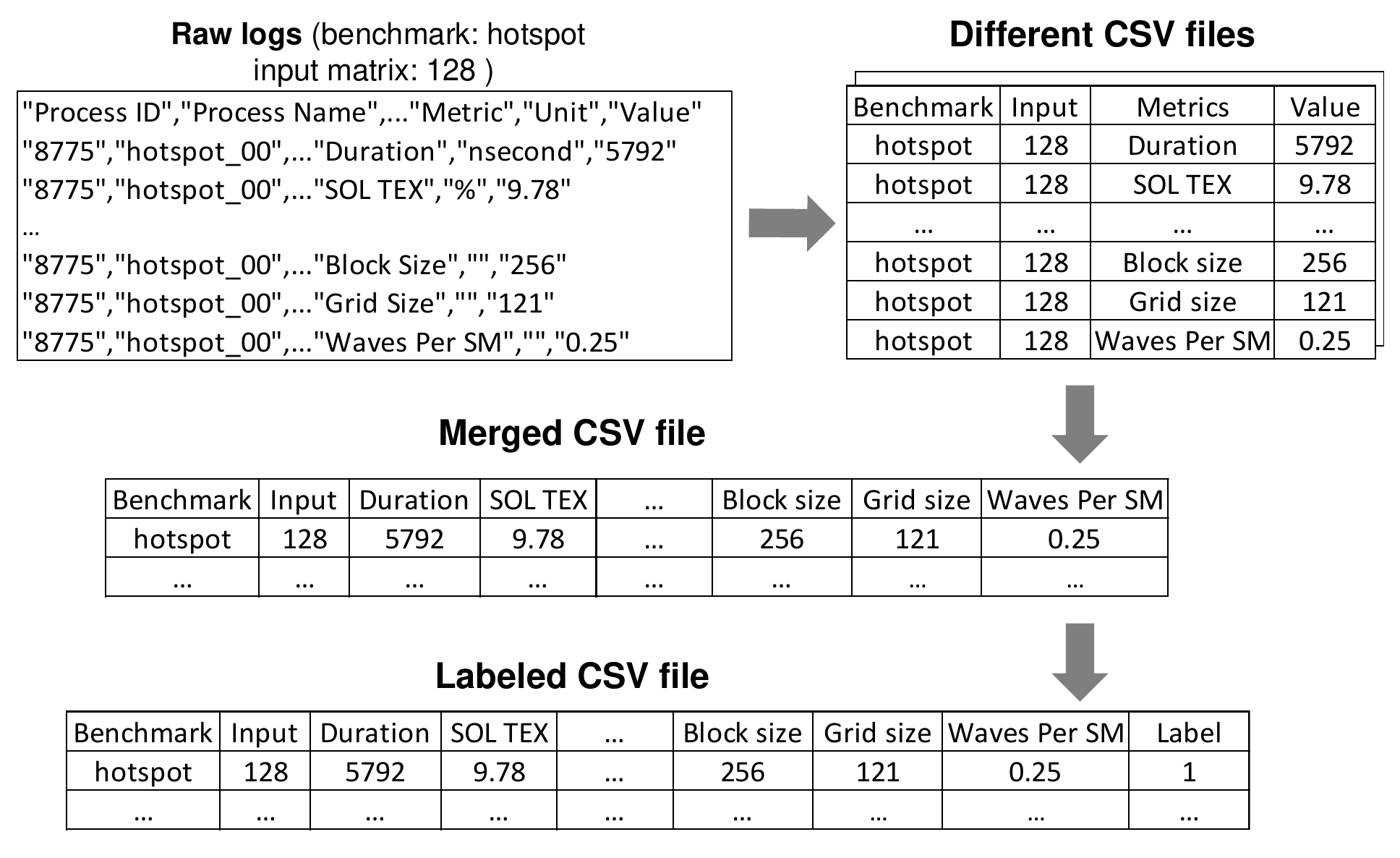}
 \caption{\textmd{XPlacer data processing pipeline}}
 \label{fig:csv}
\end{figure}

The raw data collected from XPlacer experiments is hosted at a public Github repository\footnote{https://github.com/AndrewXu22/optimal\_unified\_memory}. Figure~\ref{fig:csv} shows how the raw profiling data generated from different GPU machines are first parsed and stored into CSV files, which are merged and labeled to generate training datasets to build various machine learning models.

The readme.md of the git repo has detailed information for all released files, especially what each column means for the release CSV files. Table~\ref{tab:csv-kernel-data}, \ref{tab:csv-profiling} and \ref{tab:csv-gpu} show the labelled CSV files' columns in different sections, including those for kernels and data objects, profiling metrics, and GPU properties.  The Code variant ID column is used to represent an memory advise given to data objects in a kernel. 
\begin{table}[!htp]\centering
\caption{CSV columns for kernels and data objects}\label{tab:csv-kernel-data}
\scriptsize
\begin{tabular}{|l|l|}\toprule
\textbf{Column Name} &\textbf{Description} \\\midrule
Benchmark & Benchmark software name \\
commandLineOption& Representing data size or input files etc. \\
KernelName &GPU kernel name \\
ArrayName (Data) &Name of the array \\
ArrayID (ID) &Internal integer ID of the array \\
Variant & Code variant ID  \\
labeledVariant (label) &The best performing code variant ID \\
AllocatedDataSize (DataSize) &Memory allocation size \\
BeginAddr &Beginning address of memory allocation \\
EndAddr &Ending address of memory allocation \\
\bottomrule
\end{tabular}
\end{table}

\begin{table}[!htp]\centering
\caption{CSV columns for profiling metrics}\label{tab:csv-profiling}
\scriptsize
\begin{tabular}{|l|l|}\toprule
\textbf{Column Name} &\textbf{Description} \\\midrule
Elapsed Cycles & Cycle count of profiled code \\
TotalExecutionTime &Total execution time for this kernel.e.g \\
NumberOfCalls &Number of calls to this kernel \\
AverageExecutionTime &Average kernel execution time(ms) \\
MinExectionTime &Minimal kernel execution time \\
MaxExecutionTime &Maximal kernel execution time \\
ExecutionTimePercentage &Percentage of time spent in this kernel (\%) \\
MemoryThroughputPercentage &GPU memory throughput \\ 
SOL DRAM & DRAM utilization \\ 
SM [\%] &SM throughput \\ 
SOL L1/TEX Cache & L1/Tex cache utilization \\ 
SOL L2 Cache & L2 cache utilization \\ 
Achieved Occupancy &Achieved profiled occupancy \\
Achieved Active Warps Per SM &Profiled active warps per SM \\
CPUPageFault &CPU page fault count \\
GPUPagePault &GPU page fault count \\
HtoD &Host to Device transfer size \\
DtoH &Device to Host transfer size \\
\bottomrule
\end{tabular}
\end{table}

\begin{table}[!htp]\centering
\caption{CSV columns for GPU properties}\label{tab:csv-gpu}
\scriptsize
\begin{tabular}{|l|l|}\toprule
\textbf{Column Name} &\textbf{Description} \\\midrule
DRAM Frequency &Frequency of DRAM \\
SM Frequency &Frequency of Streaming multiprocessor \\
SM Active Cycles &Active cycle counts from SM \\
Theoretical Active Warps per SM &Theoretical Active Warps per SM \\
Theoretical Occupancy &Theoretical Occupancy \\
Block Limit SM &Max block limited by SM \\
Block Limit Registers &Max block limited by registers \\
Block Limit Shared Mem &Max block limited by shared memory \\
Block Limit Warps &Max block limited by warps \\ \midrule
\multicolumn{2}{|c|}{GPU Configuration Information}\\ \midrule
Block Size &Launch block size \\
Grid Size &Launch grid size \\
Registers Per Thread &Launch register//thread \\
Shared Memory Configuration Size &Launch shared memory configuration size \\
Static Shared Memory Per Block &Launch static shared memory \\
Threads &GPU thread count \\
Waves Per SM &Launch wave per SM \\
RemoteMap &Unified memory remote map \\
\bottomrule
\end{tabular}
\end{table}

\begin{figure}[!htbp]
 \centering
 \includegraphics[scale=0.25]{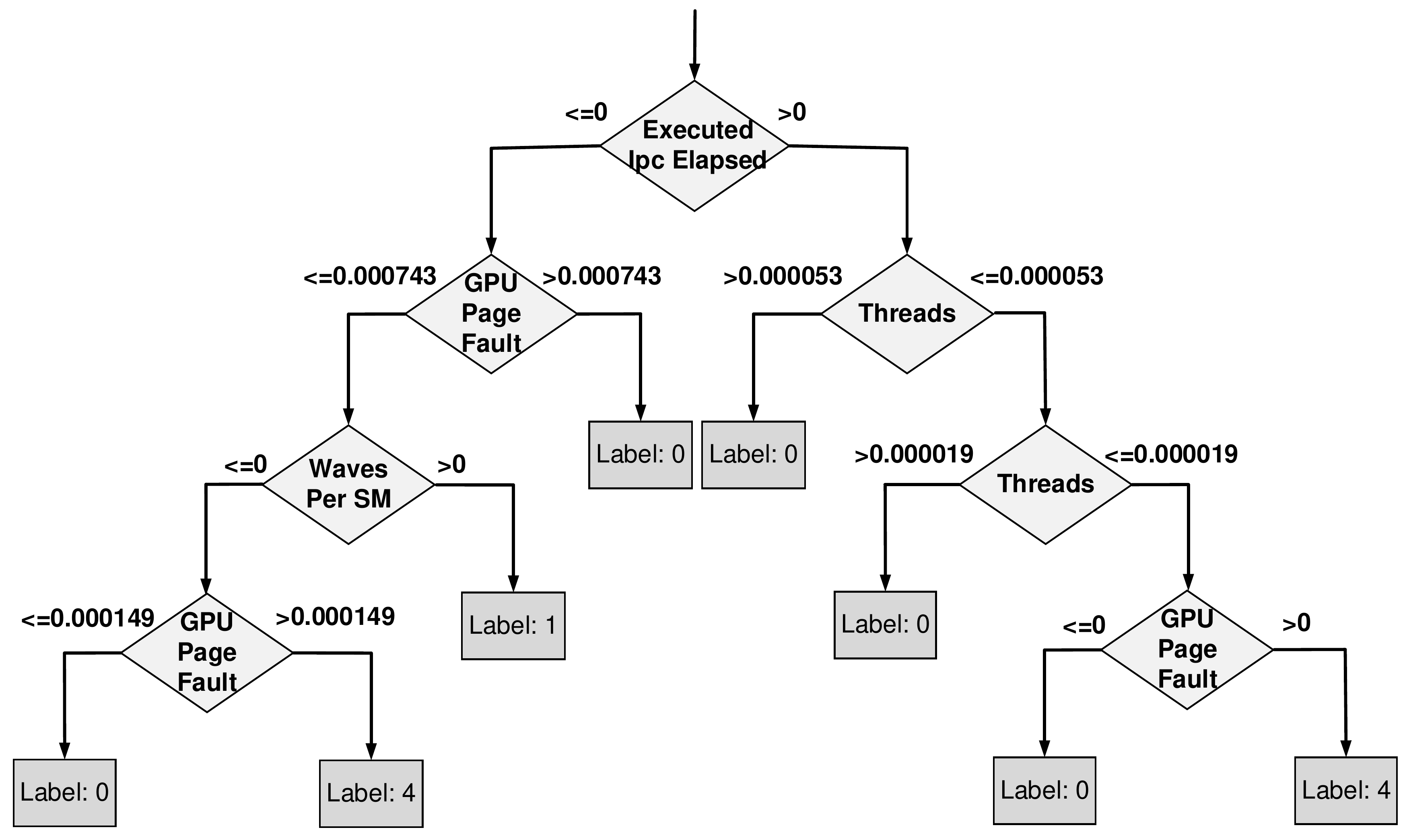}
 \caption{\textmd{A partial view of the decision tree model of XPlacer.}}
 \label{fig:decision_tree}
\end{figure}

Multiple machine learning classification models were generated by XPlacer from the training datasets. These models include Random Forest, Random Tree and Decision Tree. Figure~\ref{fig:decision_tree} shows a partial view of the decision tree model in XPlacer. 
Each tree node in the decision tree represents a decision determined by a feature with a threshold value. 
When the value of the feature is less and equal than the threshold value, the decision tree will move downward to the left node of the tree.
Otherwise, it moves downward to the right child node.
The tree traversal continues to move downward and compare various features and threshold values until it reach to a tree leaf node which represents a label value.  A label value represents the final decision made by the decision tree model.  
For Xplacer, the label value represents the memory advise that will be recommended by the machine learning model. 

\section{Experiment}
\label{results}
This section presents the FAIRification process of applying our methodology, presented in section~\ref{Fairify}, to the XPlacer datasets and the decision tree model. 

\begin{figure*}[!ht]
     \centering
    \begin{minipage}{.45\linewidth}
        \begin{subfigure}[t]{.9\linewidth}
            \includegraphics[scale=0.3]{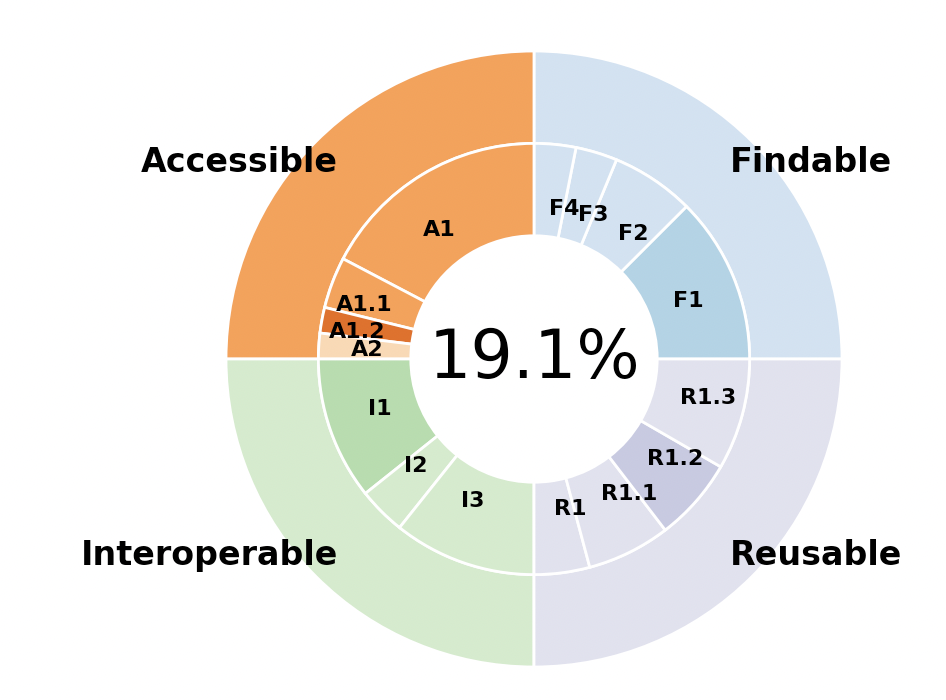}
        \end{subfigure} 
        \begin{subfigure}[b]{.9\linewidth}
            \includegraphics[scale=0.3]{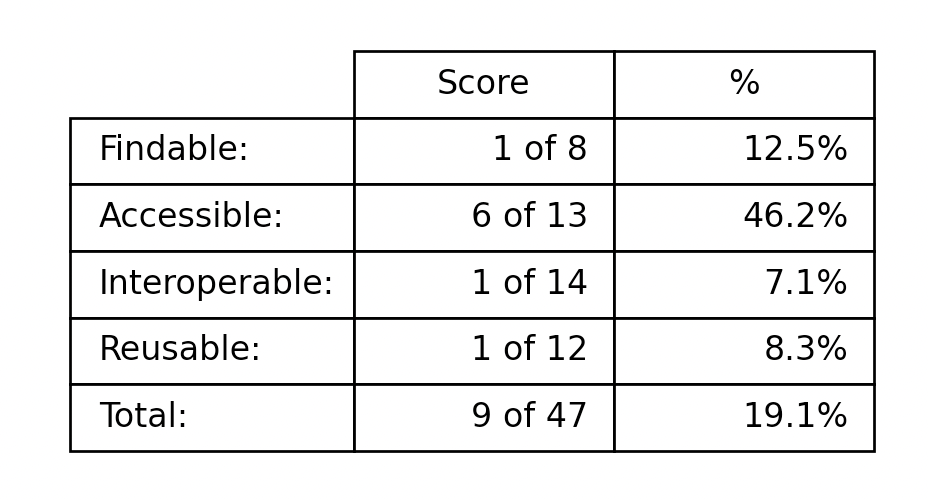}
           \caption{Initial FAIRness score}
        \label{fig:initEvalA}
        \end{subfigure} 
        
    \end{minipage}
     \begin{minipage}{.52\linewidth}
            \begin{subfigure}[t]{.9\linewidth}
                \includegraphics[scale=0.31]{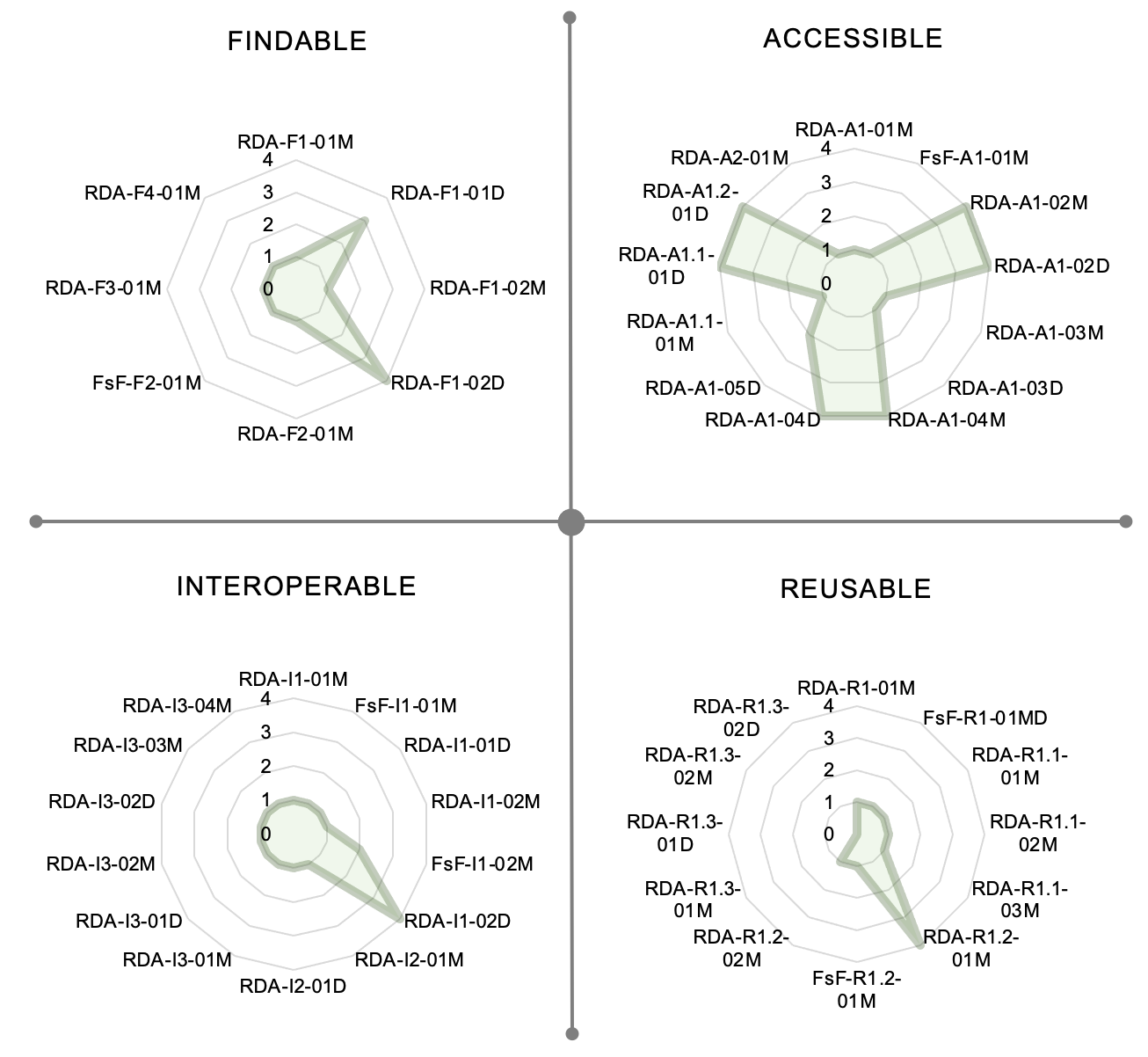}
                \caption{Initial FAIRness maturity levels}
                \label{fig:initEvalB}
            \end{subfigure}
        \end{minipage}
        \caption{Initial hybrid assessment}
        \label{fig:initEval}
        \vspace{-0.1in}
\end{figure*}

\noindent\textbf{Initial Assessment:}
The initial hybrid assessment results shown in Figure~\ref{fig:initEval}  reveal that a selected XPlacer dataset (generated on an IBM machine) has a 19.1\% FAIRness score. 
The visual reports from the RDA manual evaluation and the detailed report from the F-UJI evaluation reveal that persistent identifier is missing to fulfill the `F' FAIR principle.  
Missing license and provenance information are the major factors for its low fulfillment in the `R' FAIR principle.  
For the rest covered in `A' and `I' FAIR principles,
missing metadata information is the root cause for its low fulfillment in FAIRness.  

\noindent\textbf{Improving FAIRness:}
With the assessment feedback and recommendations, we apply the methods mentioned in Section~\ref{sec:improvingMethods} to achieve a higher assessment result.
\begin{itemize}[leftmargin=*]
\item We uploaded the XPlacer dataset to \texttt{Zenodo.org} to obtain an DOI for persistent identifier.  
With the DOI available, assessment can pass FAIR evaluation metric FsF-F1-02D.

\item As part of the uploading process, we carefully filled in required metadata information in Zenodo. The enhancement is able to fulfill eleven more metrics.

\item Basic provenance information is also provided by Zenodo. However, Zenodo does not provide formal provenance metadata such as PROV.  Therefore, only partial fulfillment is achieved for metric FsF-R1.2-01M.

\item We also chose the Creative Commons 4.0 license (CC-BY 4.0) for the XPlacer dataset to fulfill the R1.1 principle.

\item We extended the HPC ontology~\cite{liao2021hpc} to provide required rich attributes to describe fine-grain data elements. A special unit ontology (QUDT) is also used to annotate the units for numerical values to enable maximum data interoperability and fulfill the RDA-I3-01M FAIRness indicator. 

\item We used Tarql to automatically convert the corresponding CSV file into linked data using the JSON-LD format, with attributes provided by the HPC Ontology. Listing~\ref{jsonoutput} shows an example output of the conversion. A set of key-value pairs are generated to describe data cells of a CSV file. Each key is a standard metadata tag or an attribute provided by an ontology. For example, \textit{hpc:hostToDeviceTransferSize} at line 9 is used to indicate CPU to GPU data transfer size. The corresponding value at line 10 is pointing to an object defined between line 13 and 23, which include two other nested objects for precisely encoding unit (KiloByte) and value (7872.0), respectively. This level of fine-grain details is required to enable maximal interoperability among datasets. 

\item For the decision tree model,
High-level metadata describing the model is provided in the FAIRification process.  
In addition,
we leverage and extend HPC ontology to provide standard attributes to annotate fine-grain information in the decision tree.
The decision tree model can then be presented as linked data using JSON-LD format tree by annotating the tree nodes with the feature and threshold value for the non-leaf nodes and the label property for the leaf nodes.

\end{itemize}
\begin{lstlisting}[numbers=left, basicstyle=\tiny, captionpos=t, caption=Example JSON-LD output, label={jsonoutput}]
{
  "@id": "http://example.org/test.csv#L1",
  "@type": "hpc:TableRow",
  "hpc:codeVariant": "111100",
  "hpc:allocatedDataSize": 8000000,
  "hpc:arrayID": "0",
  "hpc:commandLineOption": "graph1MW.6",
  "hpc:gpuPageFault": 5,
  "hpc:hostToDeviceTransferSize": {
    "@id": "_:Nbdd222a0d12a483d8f1a4cef274f18fc"
  }
},
{
 "@id": "_:Nbdd222a0d12a483d8f1a4cef274f18fc",
  "@type": "http://qudt.org/schema/qudt/QuantityValue",
  "http://qudt.org/schema/qudt/unit": {
    "@id": "http://qudt.org/vocab/unit/KiloBYTE"
  },
  "http://qudt.org/schema/qudt/value": {
    "@type": "http://www.w3.org/2001/XMLSchema#decimal",
    "@value": "7872.0"
  }
}
\end{lstlisting}

\begin{figure*}[!ht]
     \centering
     \hspace{-0.1in}
    \begin{minipage}{.45\linewidth}
        \begin{subfigure}[t]{.9\linewidth}
            \includegraphics[scale=0.3]{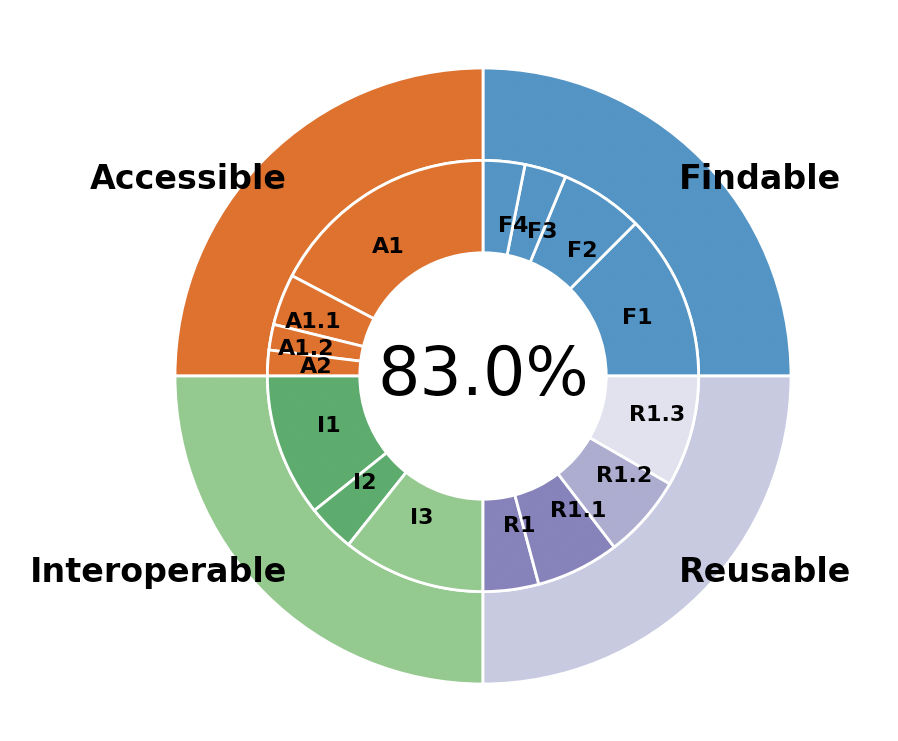}
        \end{subfigure} 
        \begin{subfigure}[b]{.9\linewidth}
            \includegraphics[scale=0.3]{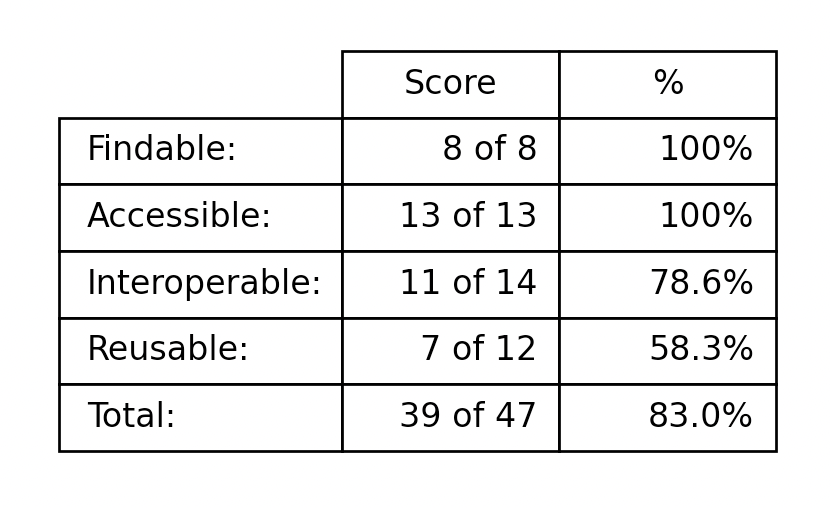}
           \caption{Final FAIRness score}
        \label{fig:finalEvalA}
        \end{subfigure} 
        
    \end{minipage}
     \hspace{-0.1in}
     \begin{minipage}{.52\linewidth}
            \begin{subfigure}[t]{.9\linewidth}
                \includegraphics[scale=0.3]{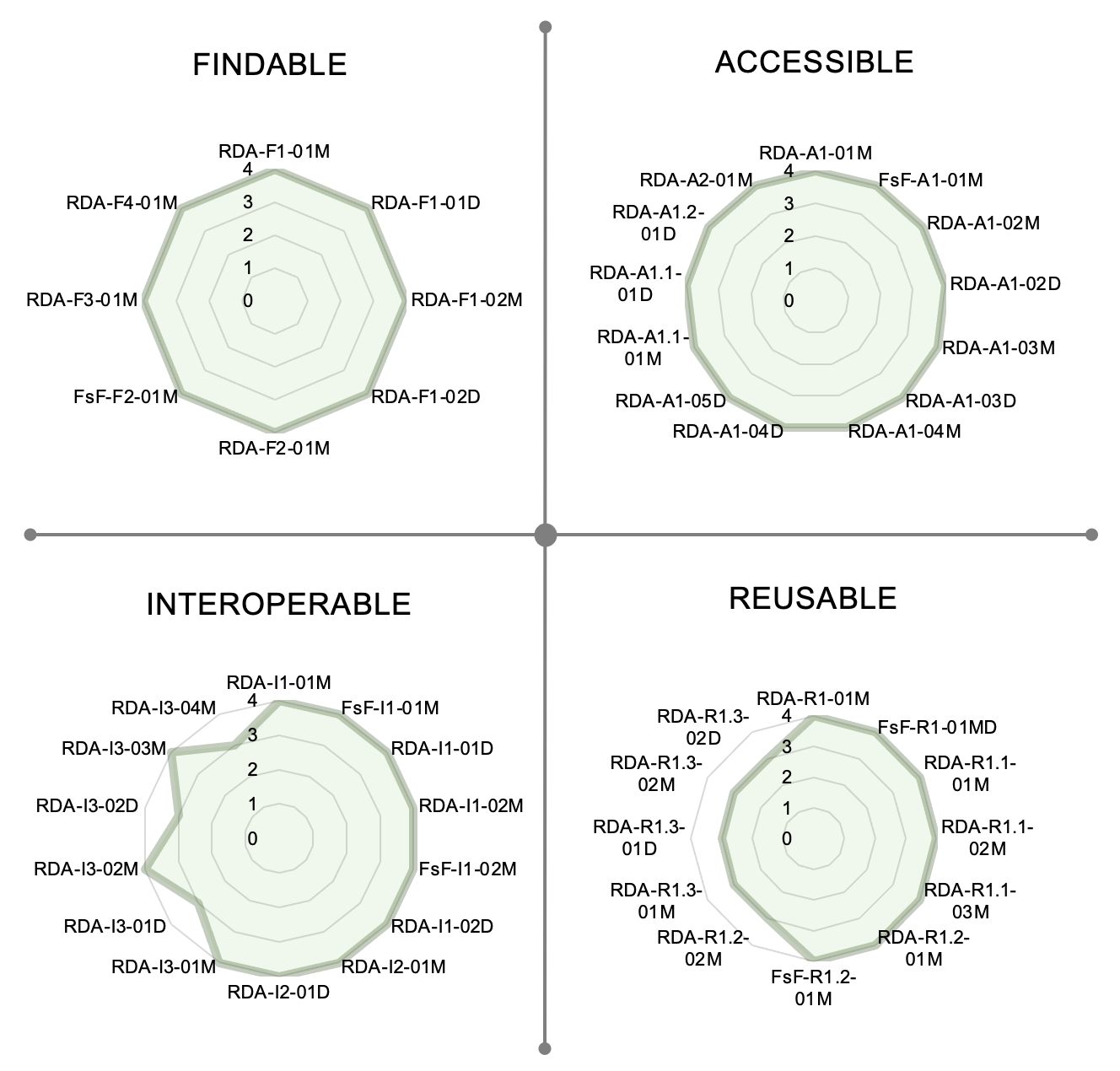}
                \caption{Final FAIRness maturity levels}
                \label{fig:finalEvalB}
            \end{subfigure}
        \end{minipage}
        \caption{Final hybrid assessment}
        \label{fig:finalEval}
        \vspace{-0.1in}
\end{figure*}

\noindent\textbf{Final FAIRness Assessment:}
The final FAIRness evaluation by the hybrid evaluation reveals a 83.0\% score after the improvements. Figure~\ref{fig:finalEval} presents the final statistics (Figure~\ref{fig:finalEvalA}) and the visual report showing the maturity level for each indicator according to the RDA manual evaluation (Figure~\ref{fig:finalEvalB}). 


The results show that the dataset does not fulfill FsF-I1-02M that requires namespaces of known semantic resources to be  present in the metadata of an object.  This is due to the limited support by the data hosting service by Zenodo to provide  namespaces for specific semantic resources.
A partial fulfillment is reported by F-UJI for FsF-R1-01MD that requires metadata for technical properties of the data file. We find that this is due to the implementation limitations on both F-UJI and Zenodo after investigation.  
The metric is considered fulfilled in our final manual assessment.
Another partial fulfillment report is for FsF-R1.2-01M that requires provenance metadata provided in a machine-readable version of PROV-O or PAV. 
We encoded the provenance information for XPlacer dataset at the fine-level granularity through HPC ontology and link to PROV-O ontology.  
However, these provenance metadata cannot be annotated into Zenodo's registry due to its limited support. This metric is also considered passed in our manual assessment.

The final assessment shows that the use of the HPC ontology significantly improves the metadata support for the XPlacer dataset, resulting in full fulfillment for all FAIRness indicators for `F and A' FAIR principles.  
However, providing full and qualified references to other data standards is still ongoing development for the HPC ontology.  This leads to a lower fulfillment, in implementation phase, for RDA-I3-01D (Data includes references to other data), RDA-I3-02D (Data includes qualified references to other data), and RDA-I3-04M (Metadata include qualified references to other data).
In the `R' FAIR principle, RDA-R1.2-02M (Metadata includes provenance information according to a cross-community languages) is also considered in implementation phase to support PROV-O ontology as the cross-community languages for provenance information.  
There is no community standard specified for data and metadata in HPC community.  We propose to use the HPC ontology as one metadata standard for HPC.
As HPC ontology is still under development, we consider RDA-R1.3-01M, RDA-R1.3-01D, RDA-R1.3-02M and RDA-R1.3-02D still in the implementation phase. 

\section{Related work}
\label{relatedwork}
There are other schemes and proposals to promote open public data and make the data easy to use by others.  
Linked Open Data (LOD) refers to all data that is published on the Web according to a set of best practices \cite{Bizer2018}.  
Berners-Lee proposed Linked Open Data (LOD) principles\footnote{https://www.w3.org/DesignIssues/LinkedData.html} with four rules to prepare the web of open-linked data. 
A 5-star scoring system is provided to evaluate if the linked open data is truly open and easy for use.  

There are various FAIRness evaluations proposed to help improving data FAIRness.
The FAIRshake was developed to promote the FAIRification of data produced by biomedical research projects and enable the establishment of community driven FAIR metrics and rubrics paired with manual and automated FAIR assessments\cite{clarke2019fairshake}. 
Authors of the FAIR Guiding Principles use 14 FAIR metrics and the web interfaces for a semi-automated FAIR evaluation service\cite{wilkinson2019evaluating}.
Many activities from different research communities start to address the demand to apply FAIR principles in their datasets. 
Cuno et al. present the FAIR evaluation of datasets for stress detection systems\cite{cuno2020fair}.   
Lamprecht and the team study the need for FAIR research software\cite{lamprecht2020towards} and suggest revised terms in FAIR principles to cover the traits in research software.  



\section{Discussion \& Future work}
\label{conclusion}
\noindent\textbf{Summary:}
 This paper presents a concrete methodology and a case study to make HPC datasets and ML models FAIR, after surveying existing techniques used to assess and improve FAIRness of scientific data. 
Our methodology can enjoy the benefits of both automatic and manual assessments while avoiding their limitations. 
It also includes a set of actionable suggestions to address reported FAIRness issues, including using ontologies to provide rich and standard data attributes. 
The experiment has shown that our methodology can effectively improve the FAIRness maturity for a selected dataset and ML model.    

\noindent\textbf{Discussion:}
To make general datasets and ML models FAIR, it can be discussed from three different aspects:
\begin{itemize}[leftmargin=*]
    \item FAIRness assessment: In this study, we have observed that existing FAIRness assessment tools have several major weaknesses:  insufficient coverage for details in FAIR principles, tendency to report biased assessment result, with emphasis to limited group of users (data curators, data stewards, and data users), and with support for only specific domains (with  attributes and vocabularies in specific scientific communities).  
    Improving the FAIRness assessment support addressing the above weaknesses can provide a trustworthy gauge of data FAIRness.  Data generators, curators, stewards and users can jointly improve the data FAIRness based on a standardized and creditable metric of data FAIRness. 
    \item FAIR-aware data store and management: Data hosting and management service have great impact to the FAIRness for the hosted/managed data. In this paper, \texttt{zenodo.org} is recommended for its support in DOI generation, general metadata, provenance and license information support.  It provides relatively smooth transition for users, who heavily rely on git repositories to store data and digital contents, to FAIRify the data.
    Several leading services for hosting/managing ML datasets and models \cite{huggingface,kaggle,hosny2019modelhub,openml} are commonly used by ML developers and users. 
    We survey and evaluate their support for FAIRness and observe several commonly seen issues: missing persistent identifier representing the datasets or models, insufficient coarse and fine-level metadata, 
    and accessing (meta)data and model is constrained by hosting service APIs.
    Providing FAIR-aware data store and management systems is also a critical factor to make general datasets and ML models FAIR. 
    \item FAIRification for ML models: As FAIR principles are applicable to any digital object, FAIRness for ML models should also be considered.
    We apply data annotation with support from HPC Ontology as an example to achieve FAIRness fulfilment for the selected decision tree model.
    However, the same approach might not be practical to many large scale ML models that contain millions or billions of parameters.
    Supporting rich metadata describing the ML models would be the alternative approach for the FAIRification.
    There is not yet a dedicated list of attributes and metadata information to represent all ML models (reflecting 'R1' sub-principle). 
    We also observe there are many information associated with the Interoperability and Reusability of a ML model cannot easily be found.
    For example, the NLP model BERT comes in many flavors: varying its size, preprocessing applied to the dataset (case and special characters), and additional fine-tuning objectives.
    Such information might not be shown by existing model hosting services and cannot be checked by any FAIRness assessment tool. 
    And last, ML models closely related to the datasets used for training.  The reusability of a ML model is likely to be limited to a specific type of dataset, or learning goal.
    Attributes describing the dataset requirements and learning goal need to be properly defined.
    Therefore, identifying the FAIR-centric metadata and attributes for the ML models (e.g.\ model architecture, training configuration, and training objectives), together with automated evaluation to assess the metadata, can greatly promote the FAIR for ML models.  
    
\end{itemize}


\noindent\textbf{Future work:}
We will continue to improve the HPC ontology to have required references to other data standards.  We hope that the HPC ontology can be adopted as one standard to FAIRify data within the HPC community. 
We will also incorporate some manual correction and assessment steps into the automated assessment step to further improve the FAIRness assessment. 
Last but not the least, we will extend our work to support FAIRification for ML models and workflows.  
Ultimately, we aim to pursue trustable machine learning by achieving better FAIRness in machine learning models and datasets.

\printbibliography

\end{document}